\documentclass[english]{IEEEtran}
\usepackage[T1]{fontenc}
\usepackage[latin9]{inputenc}
\usepackage{geometry}
\usepackage{amsmath}
\usepackage{amssymb}
\usepackage{graphicx}

\makeatletter
 
\newcommand{\noun}[1]{\textsc{#1}}
\providecommand{\tabularnewline}{\\}

\numberwithin{equation}{section}
\numberwithin{figure}{section}

\usepackage{bbold}
\usepackage[all]{xy}

\usepackage{graphicx}
\usepackage{longtable}
\usepackage{booktabs}

\usepackage{babel}

\makeatother

\usepackage{babel}
\begin{document}
\title{\noun{Identification of Stone Deterioration Patterns with Large Multimodal
Models. }}
\author{Daniele Corradetti$^{a)\,b)}$, Jos\'e Delgado Rodrigues$^{c)}$}

\maketitle
\textbf{\noun{\small{}ABSTRACT -}}\textbf{\small{} The conservation of stone-based cultural heritage sites is a critical concern for preserving cultural and historical landmarks. With the advent of Large Multimodal Models, as GPT-4omni (OpenAI), Claude 3 Opus (Anthropic) and Gemini 1.5 Pro (Google), it is becoming increasingly important to define the operational capabilities of these models. In this work, we systematically evaluate the abilities of the main foundational multimodal models to recognise and classify anomalies and deterioration patterns of the stone elements that are useful in the practice of conservation and restoration of world heritage. After defining a taxonomy of the main stone deterioration patterns and anomalies, we asked the foundational models to identify a curated selection of 354 highly representative images of stone-built heritage, offering them a careful selection of labels to choose from. The result, which varies depending on the type of pattern, allowed us to identify the strengths and weaknesses of these models in the field of heritage conservation and restoration.}{\small\par}

\section{Introduction }

The preservation of cultural heritage sites is a critical challenge
in the fields of archaeology, historic preservation, and conservation
science. Stone monuments, buildings, and artefacts are subject to
a wide range of deterioration processes over time, including weathering,
erosion, biological growth, salt crystallisation, and human-induced
damage (Doehne \& Price, 2010). Identifying and classifying these
deterioration patterns is essential for developing effective conservation
strategies and interventions. 

In recent years, advances in artificial intelligence and machine learning
have opened up new possibilities for automated analysis of stone-built
heritage. In particular, the development of large multimodal models
(LMM) - i.e., AI systems that can process and generate multiple types
of data, including text, images, and audio (Yin et al., 2023) - has
the potential to revolutionise the way we study and preserve the world\textquoteright s
stone-built heritage. Indeed, it is important to immediately note
the difference between these models and those resulting from specific
neural networks dedicated to identifying specific deterioration patterns
as they were usually developed prior to 2022. Indeed, these neural
networks were classical examples of narrow artificial intelligence
(Morris et al., 2023) visual recognition systems created ad hoc for
specific stone deterioration patterns through transfer learning from
neural networks structured for specific segmentation and classification
problems (most existing studies have focused on narrow, specialised
tasks like crack detection or material classification, often using
small datasets and custom-built algorithms cfr. Oses et al, 2014;
C\^amara et al., 2023; Mishra \& Louren\c{c}o, 2024). On the other hand,
the current LMMs, such as OpenAI\textquoteright s GPT-4omni, Anthropic\textquoteright s
Claude 3 Opus, and Google\textquoteright s Gemini 1.5 Pro, aim to
achieve a general model capable not only of recognising all specific
deterioration patterns with the same model but also of performing
complex diagnoses through linguistic operations or special cognitive
architectures and potentially suggesting specific interventions (e.g.,
Saab et al., 2024; Brown et al., 2020).

Despite the extensively studied abilities of these LMMs, the specific
skills and limitations of models for heritage conservation applications
have not been systematically evaluated. The increasing integration
of these models into human operational workflows thus necessitates
an extensive and systematic evaluation of how state-of-the-art multimodal
models perform on a diverse range of stone deterioration patterns
and anomalies, using large, representative image sets.

\begin{figure*}
\begin{centering}
\includegraphics[scale=0.2]{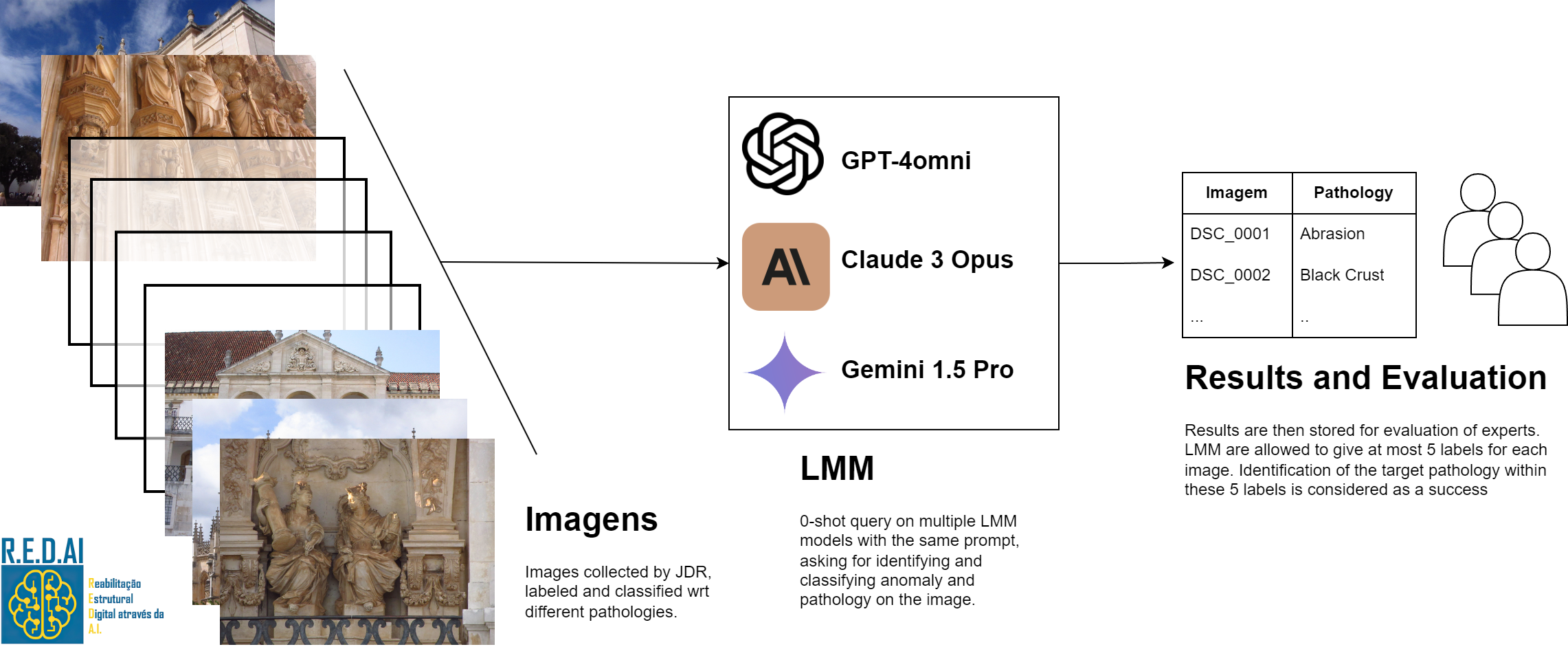}
\par\end{centering}
\caption{\label{fig:Workflow}Implemented workflow followed in this study.}
\end{figure*}

In this work, we aim to address this gap by conducting a rigorous
evaluation of three leading multimodal models - GPT-4omni (OpenAI),
Claude 3 Opus (Anthropic), and Gemini 1.5 Pro (Google) - on the task
of recognising and classifying stone deterioration patterns relevant
to world heritage conservation. We define a taxonomy of stone deterioration
patterns based on the internationally recognised literature (ICOMOS-ISCS
2008), but adapted to suit new AI interpretation tools. We then curate
a dataset of 354 high-quality images exemplifying these deterioration
patterns on stone heritage sites around the world. Using carefully
designed prompts and label sets, we assess each model\textquoteright s
performance in identifying the deterioration patterns present in each
image. Images and code for the test are available (see Section 5 Data
Availability) in order to allow replication of the test to other foundational
models or dedicated artificial cognitive entities (Fig. \ref{fig:Workflow}).

The results obtained are extremely informative, showing a significant
variation in accuracy depending on the pattern analysed. While some
patterns are identified with impressive accuracy, others seem invisible
to LMMs or indistinguishable from other patterns. A hypothesis formed
in our work group is that the lack of specific training has led large
language models (LLMs ) to adopt the common, non-technical meanings
of the words used to identify these patterns. The training of LMMs
involves the semantic association between words and images through
a process of text, video, and audio encoding into the same semantic
space, then training often uses supervised learning and contrastive
learning strategies (Radford et al., 2021). However, the absence of
a special attention to the technical meanings that certain words have
in the context of conservation of cultural heritage leads to a confusion
of terms and thus to a generic and unspecific diagnosis. A structural
problem is then the difficulty of distinguishing certain specific
patterns from mere photographs, for instance when a tactile evaluation
is required for their identification. In our group, the opportunity
to test current LMMs on such deterioration patterns was discussed.
The positive response was dictated by the need to have a realistic
overview of the possibility of using these tools profitably and integrating
them into the concrete activities of restoration and conservation
of heritage where the distinction of these patterns is still hoped
for if not required. Furthermore, we also want to stress out that
the implemented test constitutes just a first evaluation on foundational
models, without considering any finetuning or additional architectural
solution around those models. In fact, we think that the results were
encouraging, suggesting that specific training sessions on dedicated
datasets could improve drastically the results. Finally, it is worth
noting that, given the rapid evolution of the field, the aforementioned
LMMs have been trained on a large portion of the existing non-synthetic
data. Given the scarcity of original data, subsequent generations
of LMMs are increasingly oriented towards training-based on synthetic
data generated by previous versions of LMMs (Choenni et al. 2023;
Gholami et al. 2023; Li et al., 2023). While this type of training
can be very effective in teaching general skills to the models, it
is possible that a cognitive weakness in the training of the actual
generation of LMMs will likely remain influencing the successive generations,
if not properly addressed. 

\section{Evaluating the Large Multimodal Models}

A Large Multimodal Model is an AI system capable of processing and
understanding multiple modalities of information, such as text, images,
audio, and video. These models aim to mimic human-like perception
integrating the complementary nature of different data types. Recent
advancements in deep learning, particularly transformer architectures,
have enabled the development of large-scale multimodal models that
can perform a wide range of tasks across various domains (Vaswani
et al. 2017). Multimodal models learn joint representations of different
modalities (Devlin et al., 2018), allowing to capture the complex
relationships and dependencies between them. This enables the models
to perform cross-modal reasoning, generation, and retrieval tasks
(Radford et al., 2021). Even more so, in the last generation of multimodal
models, where all inputs and outputs are processed by the same neural
network, in comparison with the previous generation where the multimodality
was achieved by integrating different neural network each one trained
on a different modality (OpenAI, 2024). The versatility of LMMs has
led to their application in a wide range of domains, including healthcare,
education, entertainment, and conservation science. For our purposes,
applications in healthcare are inspirational, where LMMs have been
used for medical image analysis, disease diagnosis, and patient monitoring
(Gemini Team, 2024). While LMMs offer immense potential for automated
analysis and preservation of stone monuments, buildings, and artefacts,
it is also clear that before considering their application, a preliminary
systematic evaluation across various deterioration categories is necessary.
In this work, we decided to evaluate foundational multimodal models,
i.e., without specific training and without additional systems or
architectures, in order to recognise the native abilities of such
models. The results of this study, therefore, represent a starting
point and an evaluation of the native knowledge of the various models.
Numerous benchmarkings were inspirational, but we mainly drew inspiration
from the GPQA model (Rein et al. 2023), for which the objective of
the test is to evaluate very advanced skills through the help of experts
in the field. Given the multimodal nature of the model, the choice
of modality and evaluation method was taken into consideration. After
careful reflection and preliminary tests, we considered that the most
important and discriminating ability to evaluate would consist in
visually recognising stone deterioration patterns (Antol et al., 2015).
In fact, from our preliminary analyses, we noticed that closed-form
questions of a theoretical nature in the form of audio or text have
a very high percentage of correct answers and do not guarantee a real
applicability of the model to concrete cases. At the same time, we
estimated that visual evaluation could take place optimally through
the administration of photographic images representative of the various
patterns to the models. 

\subsection{Preparation of the test}

Deterioration patterns are the visible signs of the processes that
continually take place in any built object. By their nature, deterioration
processes are not visible and it is through the signs they impose
or originate that scientists and professionals interpret and resolve
them. In this sense, the correct observation and identification of
existing patterns is the first and most important action that conservation
professionals must take care of. Traditionally, this step is carried
out through direct inspection of the object, which involves exhaustive
observation of exposed surfaces and detailed documentation of all
relevant data collected. Having automatic processing of observational
data and correct identification of deterioration patterns would be
a relevant improvement in the preparation of conservation interventions,
which could end up being a high cost/benefit option that would contribute
to making conservation activities more affordable and attractive.
The objective of this project is to obtain an AI tool capable of correctly
identifying deterioration patterns, graphically documenting their
distribution throughout the object, and calculating areas and estimating
costs to carry out the corresponding conservation intervention. Before
embarking on designing a dedicated AI tool to specifically address
the project\textquoteright s objectives, we decided to systematically
evaluate the performance of three of the last generation LMMs on stone
deterioration patterns recognition, which required a comprehensive
benchmarking study. The preparation of the test involved two key steps:
creating a taxonomy of stone deterioration patterns and selecting
a representative set of images for each pattern. In some preliminary
trials, LMMs were allowed to use an open taxonomy, which showed that
patterns were described in colloquial terms that varied from case
to case, even when they described the same pattern. 

\subsubsection{Creating the Taxonomy}

To avoid ambiguity in interpretations as much as possible and to focus
the description made by LMMs, a comprehensive taxonomy of stone deterioration
patterns relevant to world heritage conservation has been defined.
It is largely based on the ICOMOS-ISCS Glossary of deterioration patterns
(ICOMOS 2008), with some adaptations to better meet the needs of this
activity, following the consensus gathered in a quick consultation
with some professionals in the area. After careful evaluation of the
many terms currently used to describe such patterns, the following
list was selected:
\begin{quote}
\emph{Abrasion, adherent deposit, algae, alveolisation, biological
colonisation, black crust, blistering, chipping, contour spalling,
corrosion of inserted elements, crack, craquele, dark diffuse biocolonisation,
deformation, degraded joint filling, detachment of mortar layer, differential
erosion, discolouration, efflorescence, encrustation, erosion, film,
flaking, fracture, fragmentation, gap, graffiti, granular disintegration,
lichens, loose deposit, misalignment elements, moist area, moss, open
joint, patina, perforation, pitting, plant, powdering, soiling, spalling,
staining, sugaring, thin black deposit, unaesthetic joint filling,
unaesthetic patch repair}.
\end{quote}
\begin{center}
\begin{figure*}
\centering{}\includegraphics[scale=0.85]{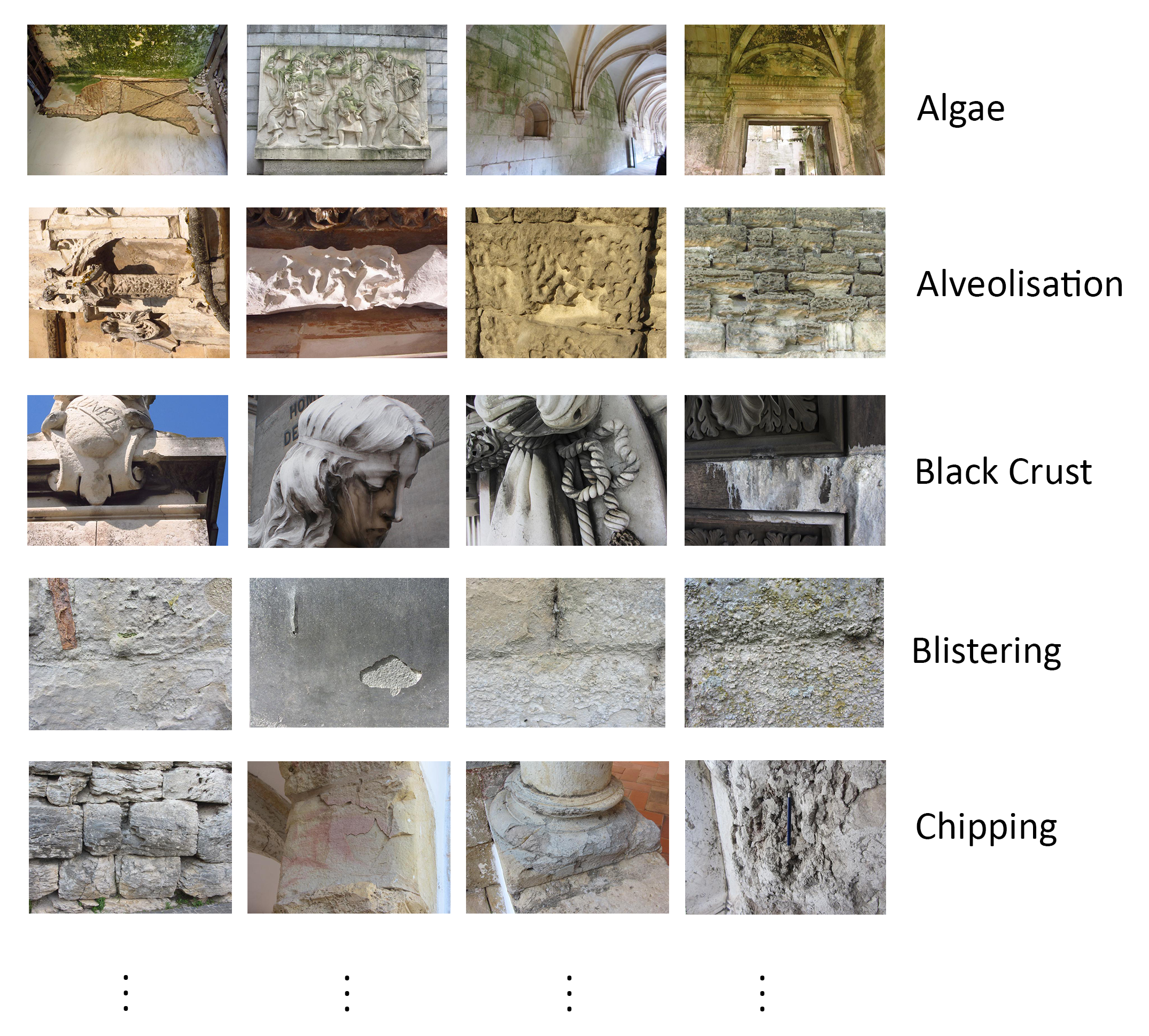}\caption{\label{fig:Example-of-the}Example of the deterioration patterns used
for benchmarking }
\end{figure*}
\par\end{center}

Fig. \ref{fig:Example-of-the} illustrates some of the deterioration
patterns used to test the models in the present study. This list covers
a wide range of stone deterioration patterns, including the major
categories of damage signs commonly encountered in stone-built objects:
patterns linked to visual disturbances; patterns representing erosion
or mass losses; patterns with direct or indirect connection with structural
stability issues. As discussed later in this article, LMMs showed
great difficulties in identifying some chosen patterns, but no effort
was made here to adapt or change the number and type of patterns to
see how such adaptations could impact the capabilities of the tested
models. However, such adaptations are not to be excluded in future
tests if deemed useful to benefit the models learning capacity and
to improve the produced outputs. 

\subsubsection{Selecting the Images}

The next step in preparing the test was to curate a dataset of high-quality
images representative of each stone deterioration pattern in our taxonomy.
We started with a large archive of over 8,000 images collected from
various sources, including field surveys and conservation reports.
The images depict stone heritage sites from around the world, spanning
different historical periods, architectural styles, and geological
contexts. Although they cover a wide variety of patterns, the images
were not taken with this study in view and are therefore certainly
not the ideal set that one could hope to test. To select the most
suitable images for our testing, we established a set of criteria.
First, each image had to clearly contain the \textquotedblleft Target
Pattern\textquotedblright{} as a major occurrence in its contents.
Second, the images had to be of sufficient resolution and quality
to allow for detailed analysis by the LMMs. Third, we aimed to include
a diverse range of stone types, surface textures, and environmental
conditions to assess the models\textquoteright{} robustness and generalisation
capabilities. It is worth noting that while each image was selected
with a specific Target Pattern in mind, they always contain other
patterns that the models were asked to identify as well.

\subsection{Selection of the models}

For this study, we selected three state-of-the-art foundational multimodal
models released in 2024: OpenAI GPT-4omni (OpenAI, 2024), Anthropic
Claude 3 Opus (Anthropic, 2024), and Google Gemini 1.5 Pro (Google,
2024). These models were chosen based on their very strong performance
on all general benchmarks (e.g., MMLU, GPQA, ARC-Challenge, etc.)
and their ability to handle a wide range of tasks across different
domains. OpenAI GPT-4omni is an extension of the GPT (Generative Pre-trained
Transformer) architecture, which has shown remarkable success in natural
language processing tasks. GPT-4omni incorporates visual processing
capabilities, enabling it to understand and generate both text and
images. The model has been trained on a vast amount of web-scale data,
processes multiple data types simultaneously, thus improving its ability
to understand and generate diverse outputs. Having real-time responses,
being scalable and an improved efficiency, this model is naturally
one of the best candidates for deployment of real applications to
cultural heritage conservation tasks. Claude 3 Opus is Anthropic\textquoteright s
next generation model offering enhanced capabilities in multimodal
processing. Claude 3 Opus builds upon the successes of the original
Claude model, which demonstrated strong performance on language understanding
and generation tasks. Moreover, Anthropic is known for its focus on
ethical and robust AI systems. As GTP4-omni, Claude 3 Opus processes
various visual formats, making it potentially suitable for detailed
heritage site analysis. Google Gemini 1.5 Pro is the most advanced
model of Google\textquoteright s Gemini series of multimodal models.
It shows excellent results on a vast range of tasks, including image
captioning, visual question answering, and image-text retrieval. Being
released by Google, Gemini 1.5 Pro is probably the easiest model to
scale and the most cost-effective, which thus makes it one of the
best candidates for deploying robust solutions with possibly high-volume
of inference requests. Indeed, Gemini 1.5 Pro has a context window
of 1 million token, which would translate in a 40 minutes long video,
the possible result of a drone inspection of a site (Google, 2024).

\subsection{Implementation of the Test}

To evaluate the selected multimodal models on the task of stone deterioration
patterns recognition, we implemented an automatised testing pipeline
(Fig. \ref{fig:Workflow}) which included a simple preprocessing and
a prompt engineering feature. The first step in the testing pipeline
was to preprocess the curated dataset of images in a way compatible
with the input size of the selected models. To get accurate and concise
responses from the models, we used a prompt that allowed the model
to provide as many patterns as necessary up to five, but only using
the list of stone patterns defined in our taxonomy. We then utilized
the official APIs provided by OpenAI, Anthropic, and Google to interact
with their respective multimodal models. The model\textquoteright s
response, containing the identified patterns, was then received and
processed. Finally, the image filename, true pattern label, and the
model\textquoteright s identified patterns were stored in a results
list for the performance evaluation. 

\section{Results }

\begin{figure*}
\begin{centering}
\includegraphics[scale=0.83]{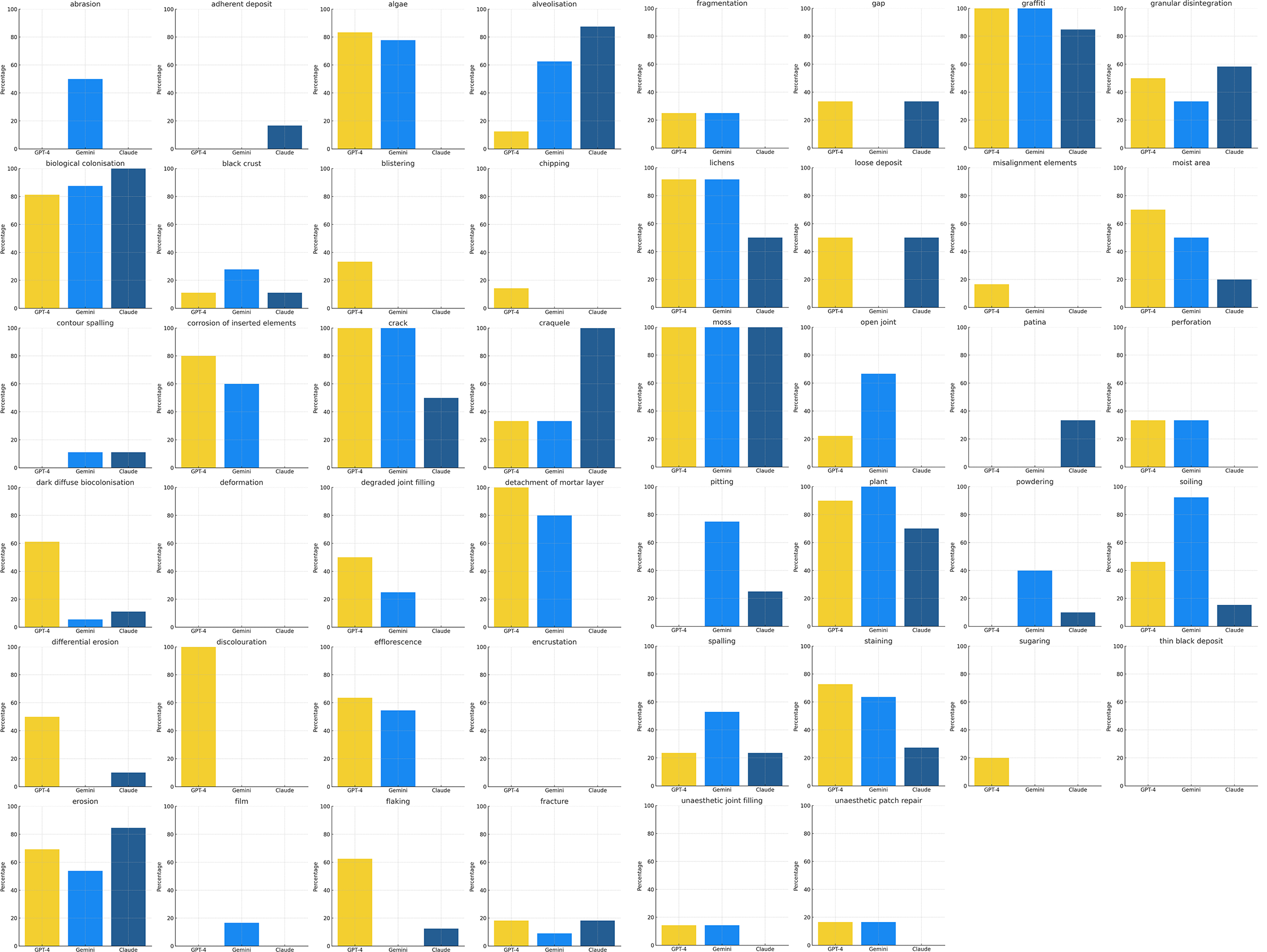}
\par\end{centering}
\caption{\label{fig:Rate-of-successTarget}Rate of success of each model on
the ``Target deterioration pattern''. \emph{Yellow} is GPT-4omni,
\emph{light blue} Gemini 1.5 Pro and \emph{blue} is Claude 3 Opus}
\end{figure*}

After the model response, the results were collected and then manually
validated. The first action was to confirm whether or not the \textquotedblleft target
pattern\textquotedblright{} was one of the identified patterns. Then,
the original image was opened to check which of the identified patterns
were actually present in the image. The first relevant result is expressed
in the rate of success the models showed in identifying the \textquotedblleft Target
Pattern\textquotedblright{} (Table \ref{tab:Success-rates-on} and
Fig. \ref{fig:Rate-of-successTarget}). Indeed, it is important for
us to be sure that the most evident or relevant deterioration pattern
is systematically identified by the LMMs.
\begin{figure*}
\begin{centering}
\includegraphics[scale=0.75]{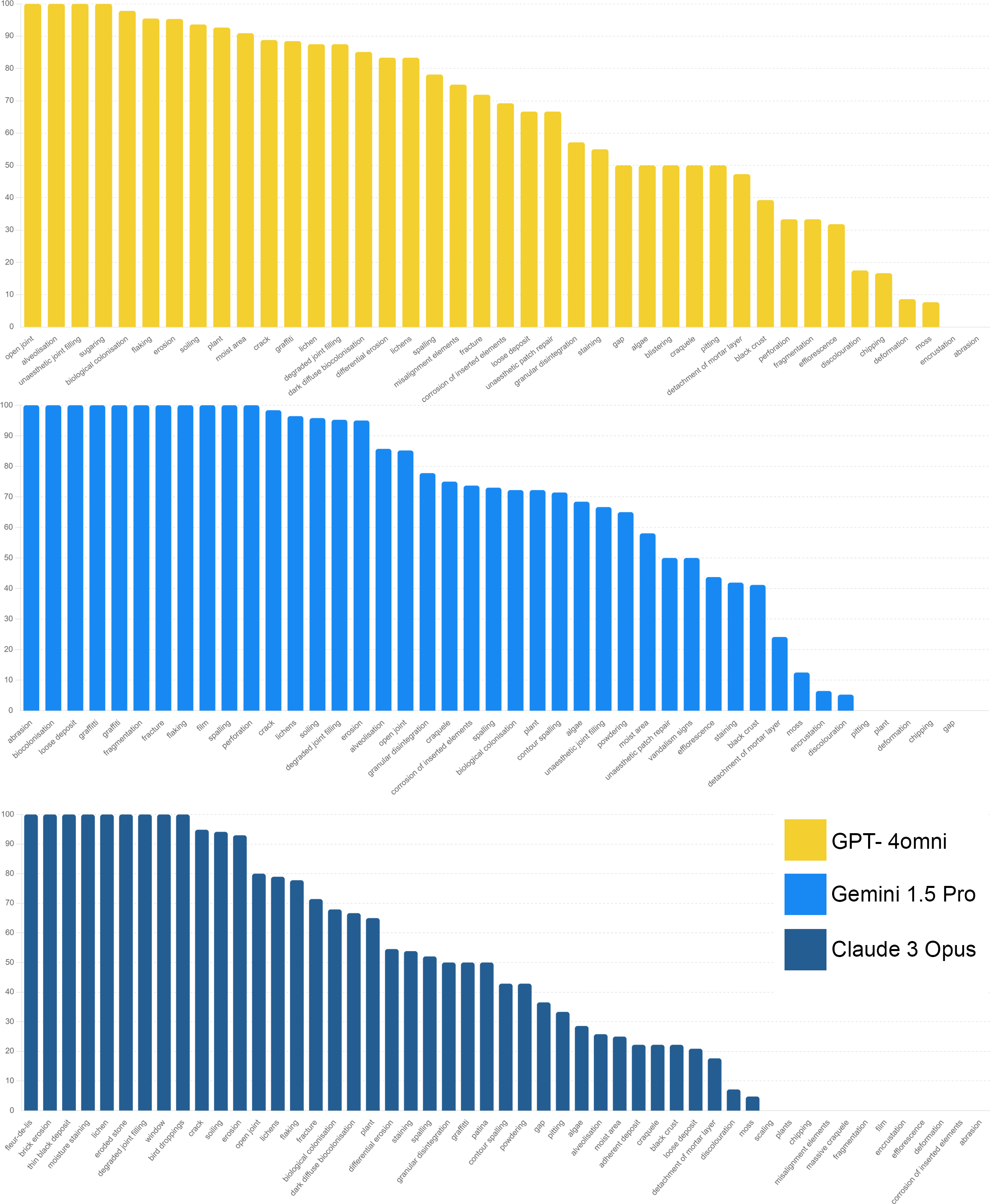}
\par\end{centering}
\caption{\label{fig:Rate-of-successIdentifying}Success rate in the identification
of the presence of a deterioration pattern (as openly chosen by the
models). In the image the color \emph{yellow} represents the model
GPT-4omni, the color \emph{light blue} represents Gemini 1.5 Pro and
the \emph{blue} is for Claude 3 Opus.}
\end{figure*}
\begin{table*}
\bigskip{}

\begin{centering}
\begin{tabular}{cccc}
\multicolumn{1}{c|}{\emph{Model}} & \multicolumn{1}{c|}{\emph{Targets correctly identified}} & \multicolumn{1}{c|}{\emph{Total possible}} & \emph{Success rate (\%)}\tabularnewline
\hline 
\textbf{GPT-4omni} & \textbf{149} & 354 & \textbf{42.1\%}\tabularnewline
Gemini 1.5 Pro & 138 & 354 & 39\%\tabularnewline
Claude 3 Opus & 86 & 354 & 24.3\%\tabularnewline
\end{tabular}
\par\end{centering}
\bigskip{}

\caption{\label{tab:Success-rates-on}Success rates on the identification of
the target deterioration pattern}
\end{table*}
\begin{table*}
\bigskip{}

\begin{centering}
\begin{tabular}{cccc}
\multicolumn{1}{c|}{\emph{Model}} & \multicolumn{1}{c|}{\emph{Deterioration patterns correctly identified}} & \multicolumn{1}{c|}{\emph{Total identification}} & \emph{Success rate (\%)}\tabularnewline
\hline 
GPT-4omni & 677 & 1031 & 65.6\%\tabularnewline
\textbf{Gemini 1.5 Pro} & \textbf{741} & \textbf{1066} & \textbf{69.5\%}\tabularnewline
Claude 3 Opus & 745 & 1263 & 58.9\%\tabularnewline
\end{tabular}
\par\end{centering}
\caption{\label{tab:Success-rates-inOpenly}Success rates in the identification
of openly chosen patterns}
\end{table*}

Besides the capability to identify the target patterns, it was also
considered as relevant to know the rate of success in identifying
non-target patterns present in the images, as openly taken by the
model from the prescribed list of patterns (Table \ref{tab:Success-rates-inOpenly}
and Fig. \ref{fig:Rate-of-successIdentifying}). 
\begin{figure*}
\begin{centering}
\includegraphics[scale=0.8]{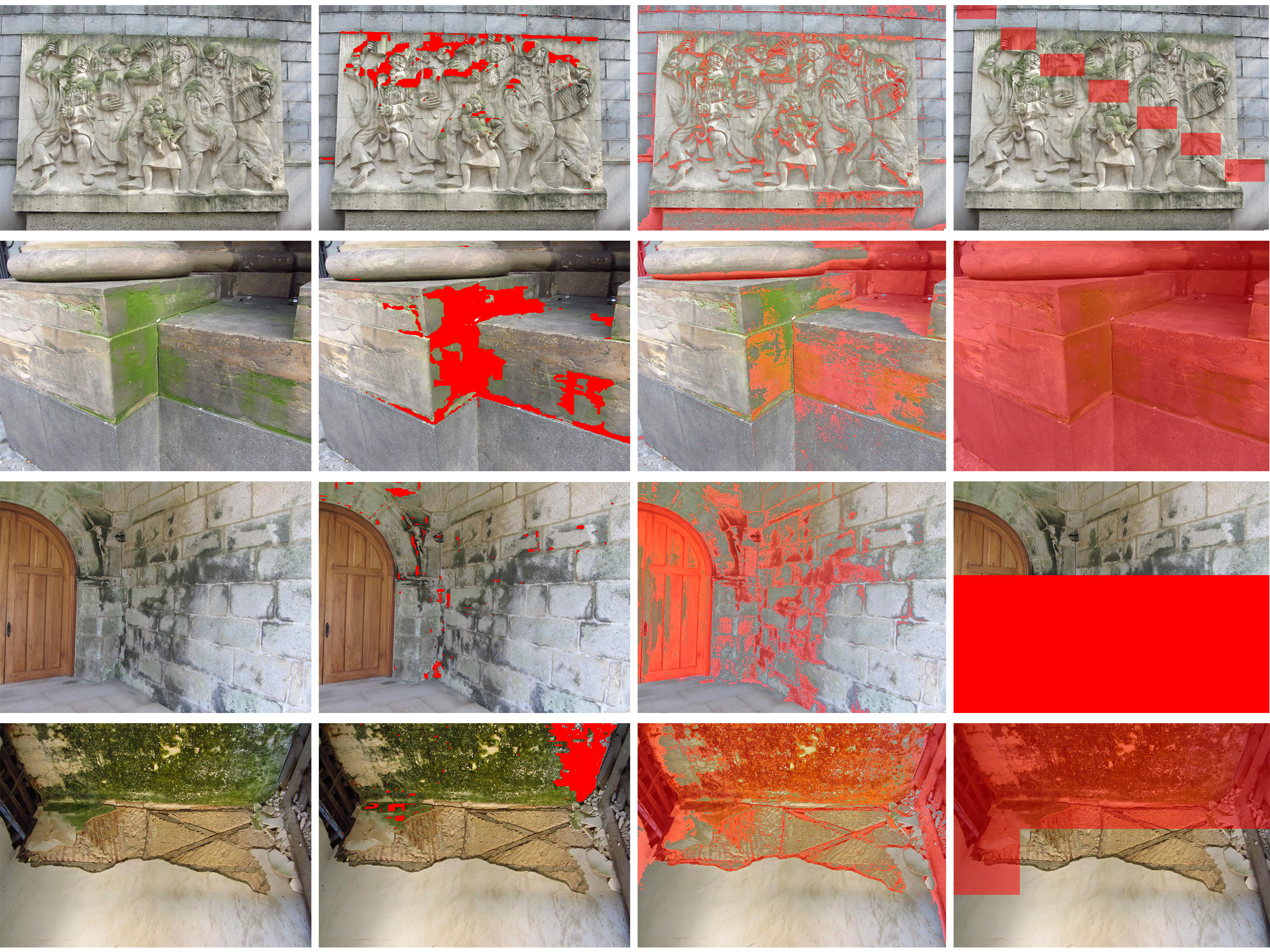}
\par\end{centering}
\caption{\label{fig:Colour}Example of additional queries to visualise the
answers reproducibility.}
\end{figure*}

\subsection{Target Patterns}

In this test results show (Table \ref{tab:Success-rates-on}) that
the best performing model was that from OpenAI, i.e., the GPT-4omni
model, which identified the target deterioration pattern in 42.1\%
of the cases, followed by Gemini 1.5 Pro, which scored a 38.9\% rate
of success, while Claude 3 Opus scored only 24.3\%.

Results also show (Fig. \ref{fig:Rate-of-successIdentifying}) that
GPT-4omni, despite its overall best performance, was not the best
performing model across all target patterns. Indeed, Gemini 1.5 Pro
scored way better than GPT-4omni in identifying numerous patterns
(e.g. \textquotedblleft abrasion\textquotedblright , \textquotedblleft alveolisation\textquotedblright ,
\textquotedblleft open joint\textquotedblright , etc.) and even Claude
3 Opus, the least performing, out passed it in identifying \textquotedblleft craquele\textquotedblright{}
and \textquotedblleft alveolisation\textquotedblright . To a certain
extent, this can be taken as an indication of the lack of reliability
of the models for this specific use.

As a general rule of thumb, we saw that LMMs are better in the spontaneous
identification of patterns in an open way than in identifying the
target pattern allocated to each image (Table 2). In a simple assessment,
this discrepancy seems to be explicable by the fact that in reality
they answer two distinct questions. The target pattern answers to
this implicit question: \textquotedblleft is this specific pattern
present?\textquotedblright , while the open identification answers
a different one: \textquotedblleft are any of these 46 patterns present?\textquotedblright .
The first set of answers in Table \ref{fig:Rate-of-successTarget}
suggests that the models are still far from being reliable and that
they fall short in guaranteeing that they are able to identify any
specific deterioration pattern. On the other hand, the results in
Table 2 show that the models already have a vast vocabulary and identification
processes incorporated, although they still seem not sufficiently
structured and robust, meaning that, for the most part, they are unusable
for practical purposes.

Overall, these first results mean that existing models, despite their
enormous capabilities, are not prepared to solve problems in the practice
of conservation and restoration of built heritage. In fact, for any
image identified as a paradigm of a given pattern of deterioration,
only when the success rate falls very close to 100\% will it be considered
ready to be used for practical purposes. 

Besides the rate of success in identifying the target pattern, it
is also of interest the models\textquoteright{} overall success rates
in identifying the presence of patterns freely taken from the open
list. In this case Gemini 1.5 Pro scored a better result, with 11
patterns accurately chosen against 5 from GPT-4omni. Claude 3 Opus
has 9 accurately chosen patterns, but 6 of them were not in the short
list of allowed patterns, for instance \textquotedblleft window\textquotedblright ,
\textquotedblleft eroded bricks\textquotedblright , etc. In strict
terms, they could be taken as erroneous identifications, but we opted
to leave them as produced to illustrate this peculiar behaviour of
the model.

\subsection{Interpretability}

An important element in developing and deploying reliable and robust
AI models for the conservation and restoration of stone involves understanding
the consistency and reliability in diagnostic capabilities. This is
particularly true for LMMs, whose results are not deterministic but
have a certain degree of variability depending on the \textquoteright temperature\textquoteright{}
at which the model is queried. To better understand the process of
pattern recognition and its robustness, we asked GPT-4omni to highlight
in red the areas affected by the pattern for which the model had obtained
a high identification score, i.e., the presence of algae in the illustrated
example (Fig. \ref{fig:Colour}). In this figure, the first image
is the original, part of the test, while the other three are the results
obtained in three separate sessions with the same prompt, i.e.,
\begin{quote}
\texttt{\textbf{prompt}}\texttt{='''Colour in red the parts that are
subject to algae pathology'''.}
\end{quote}
As can be seen, the model is not consistent for the same prompt, providing
very different results. On one hand, this is encouraging, suggesting
that there is significant room for improvement by providing the model
with specific fine-tuning, precise instructions, and diagnostic guardrails.
On the other hand, our analysis highlights the need for the construction
of a specific cognitive architecture, in the absence of which the
results of the foundational model appear to be of weak consistency
and thus of questionable reproducibility.

\section{Conclusions and Future Developments}

The study carried out with a selection of deterioration patterns typically
found in built heritage objects allowed us to verify that the LMMs
tested here were not specifically trained for the conservation and
restoration environment of built heritage. Despite the enormous capabilities
that can be attributed to them, the success rates in identifying patterns
are still far from what will be required of them as working instruments
to produce usable results. The three LMMs tested have a broad vocabulary
relevant to the area, but lack the \textquotedblleft understanding\textquotedblright{}
of concepts and terminology specific to the conservation and restoration
domain. Therefore, a specific improvement process will be necessary
to bring them to a level of proficiency compatible with the needs
of professionals in this field of activity. 

\section{Data Avalaibility}

All data, test images, code and results are available at the url:
\texttt{GitHub.com/DCorradetti/REDAI\_Id\_Pattern}

\section{Acknowledgements}

This study was carried out under the REDAI project and the Authors
thank all team REDAI for the input and fruitful discussions but most
of all Jos\'e Paulo Costa for the conceptualisation of the project.

\section{References}
\begin{itemize}
\item Anthropic. (2024, March 4). \emph{Introducing the next generation
of Claude}. url: https://www.anthropic.com/news/claude-3-family, retrieved
May 27, 2024.
\item Antol, S., Agrawal, A., Lu, J., Mitchell, M., Batra, D., Zitnick,
C. L., \& Parikh, D. (2015). \emph{VQA: Visual question answering}.
In Proceedings of the IEEE International Conference on Computer Vision
(pp. 2425-2433). https://doi.org/10.1109/ICCV.2015.279
\item Brown, T. B., Mann, B., Ryder, N., Subbiah, M., Kaplan, J. D., Dhariwal,
P., ... \& Amodei, D. (2020). \emph{Language models are few-shot learners}.
arXiv preprint arXiv:2005.14165.
\item Camara, A., de Almeida, A., Ca\c{c}ador, D., \& Oliveira, J. (2023). \emph{Automated
methods for image detection of cultural heritage: Overviews and perspectives}.
Archaeological Prospection, 30(2), 153--169. https://doi.org/10.
1002/arp.1883
\item Choenni S. , Busker T. \& Bargh M. S. , \emph{Generating Synthetic
Data from Large Language Models}, 2023 15th International Conference
on Innovations in Information Technology (IIT), Al Ain, United Arab
Emirates, 2023, pp. 73-78, doi: 10.1109/IIT59782.2023.10366424.
\item Devlin, J., Chang, M. W., Lee, K., \& Toutanova, K. (2018). \emph{Bert:
Pre-training of deep bidirectional transformers for language understanding}.
arXiv preprint arXiv:1810.04805.
\item Doehne, E., \& Price, C. A. (2010). \emph{Stone conservation: An overview
of current research}. Getty Publications.
\item Gemini Team, Google. (2024). \emph{Advancing multimodal medical capabilities
of Gemini}. arXiv preprint arXiv:2405.03162. https://arxiv.org/pdf/2405.03162
\item Gholami, S., \& Omar, M. (2023). \emph{Does Synthetic Data Make Large
Language Models More Efficient?}. ArXiv, abs/2310.07830. https://doi.org/10.48550/arXiv.2310.07830.
\item Google. (2024, February 15). \emph{Introducing Gemini 1.5, Google's
next-generation AI model}. https://blog.google/technology/ai/google-gemini-next-generation-model-february-2024/
retrieved May 27, 2024.
\item Li, Z., Zhu, H., Lu, Z., \& Yin, M. (2023). \emph{Synthetic Data Generation
with Large Language Models for Text Classification: Potential and
Limitations}. ArXiv, abs/2310.07849
\item Mishra, M., \& Louren\c{c}o, P. B. (2024). \emph{Artificial intelligence-assisted
visual inspection for cultural heritage: State-of-the-art review}.
Journal of Cultural Heritage, 66, 536-550. https://doi.org/10.1016/j.culher.2024.01.005
\item Morris, M. R., Sohl-dickstein, J., Fiedel, N., Warkentin, T., Dafoe,
A., Faust, A., ... \& Legg, S. (2023). \emph{Levels of AGI: Operationalizing
Progress on the Path to AGI}. arXiv preprint arXiv:2311.02462.
\item OpenAI (2024, May 13) \emph{Hello GPT-4o, }url: https://openai.com/index/hello-gpt-4o/
retrieved May 27, 2024. 
\item ICOMOS-ISCS. (2008). \emph{Illustrated glossary on stone deterioration
patterns}. ICOMOS International Scientific Committee for Stone (ISCS).
\item Oses, Noelia; Dornaika, Fadi; Moujahid, Abdelmalik (2014). \emph{Image-Based
Delineation and Classification of Built Heritage Masonry}. Remote
Sensing, 6(3), 1863--1889. doi:10.3390/rs6031863 
\item Radford, A., Kim, J. W., Hallacy, C., Ramesh, A., Goh, G., Agarwal,
S., Sastry, G., Askell, A., Mishkin, P., Clark, J., Krueger, G., \&
Sutskever, I. (2021). \emph{Learning Transferable Visual Models From
Natural Language Supervision}. In International Conference on Machine
Learning. https://api.semanticscholar.org/CorpusID:231591445
\item Rein, D., Hou, B. L., Stickland, A. C., Petty, J., Pang, R. Y., Dirani,
J., Michael, J., \& Bowman, S. R. (2023). \emph{GPQA: A Graduate-Level
Google-Proof Q\&A Benchmark}. arXiv preprint arXiv:2311.12022. https://arxiv.org/abs/2311.12022
\item Saab, K., Tu, T., Weng, W.-H., Tanno, R., Stutz, D., Wulczyn, E.,
Zhang, F., Strother, T., Park, C., Vedadi, E., Chaves, J. Z., Hu,
S.-Y., Schaekermann, M., Kamath, A., Cheng, Y., Barrett, D. G. T.,
Cheung, C., Mustafa, B., Palepu, A., ... Natarajan, V. (2024). \emph{Capabilities
of Gemini Models in Medicine}. arXiv. https://arxiv.org/abs/2404.18416
\item S\'anchez-Aparicio, L. J., Del Pozo, S., Ramos, L. F., Arce, A., \&
Fernandes, F. M. (2018). \emph{Heritage site preservation with combined
radiometric and geometric analysis of TLS data}. Automation in Construction,
85, 24-39. https://doi.org/10.1016/j.autcon.2017.09.023
\item S\'anchez-Aparicio, L. J., Masciotta, M. G., García-Alvarez, J., Ramos,
L. F., Oliveira, D. V., Martín-Jim\'enez, J. A., ... \& Monteiro, P.
(2020). \emph{A digital twin approach towards the assessment of seismic
vulnerability of historic masonry buildings}. In C. Calderini, M.
G. Masciotta, \& L. Mazzolani (Eds.), \emph{12th International Conference
on Structural Analysis of Historical Constructions} (pp. 1743-1761).
https://doi.org/10.1007/978-3-030-60796-5\_139 
\item Vaswani, A., Shazeer, N., Parmar, N., Uszkoreit, J., Jones, L., Gomez,
A. N., ... \& Polosukhin, I. (2017). \emph{Attention is all you need}.
In Advances in neural information processing systems (pp. 5998-6008).
\item Yin S., Fu C., Zhao S., Li K., Sun X., Xu T., Chen E. (2023) \emph{A
Survey on Multimodal Large Language Models} arXiv:2306.13549 https://doi.org/10.48550/arXiv.2306.13549 
\end{itemize}
\bigskip{}
a) Departamento de Matem\'atica,

Universidade do Algarve, 

Campus de Gambelas, 

8005-139 Faro, Portugal

a55499@ualg.pt\\
b) Grupo de F\'isica Matem\'atica

Instituto Superior T\'ecnico

Av. Rovisco Pais

1049-001 

Lisboa Portugal\\
c) Consultant in Conservation of Cultural Heritage 

j.delgado.rodrigues@gmail.com
\end{document}